\pdfoutput=1
\documentclass[11pt]{article}

\usepackage[final]{ACL2023}

\usepackage{times}
\usepackage{latexsym}
\usepackage{amsmath}
\usepackage[T1]{fontenc}
\usepackage{graphicx}

\usepackage[utf8]{inputenc}

\usepackage{microtype}

\usepackage{inconsolata}
\usepackage{caption}
\usepackage{booktabs}
\title{"Beware of deception": Detecting Half-Truth and Debunking it through Controlled Claim Editing}

\author{Singamsetty Sandeep \\ IIT Bombay  \\ \texttt{sandysingam@cse.iitb.ac.in}
        \And
        Nishtha Madaan \\ IBM Research \\ \texttt{nishthamadaan@in.ibm.com}
        \AND
        Sameep Mehta \\ IBM Research \\ \texttt{sameepmehta@in.ibm.com}
        \And
        Varad Bhatnagar \\ IBM Research \\ \texttt{varad.bhatnagar@ibm.com} 
        \And
        Pushpak Bhattacharyya \\ IIT Bombay \\ \texttt{pb@cse.iitb.ac.in}      
        }
\begin{document}
\maketitle
\begin{abstract}
The prevalence of \textit{\textbf{half-truths}}, which are statements containing some truth but that are ultimately deceptive, has risen with the increasing use of the internet. To help combat this problem, we have created a comprehensive pipeline consisting of a half-truth detection model and a claim editing model. Our approach utilizes the T5 model for controlled claim editing; \textit{"controlled"} here means precise adjustments to select parts of a claim. Our methodology achieves an average BLEU score of \textbf{0.88} (on a scale of 0-1) and a \textit{disinfo-debunk} score of \textbf{85\%} on edited claims. Significantly, our T5-based approach outperforms other Language Models such as GPT2, RoBERTa, PEGASUS, and Tailor, with average improvements of 82\%, 57\%, 42\%, and 23\% in \textit{disinfo-debunk} scores, respectively. By extending the LIAR-PLUS dataset, we achieve an F1 score of \textbf{82\%} for the half-truth detection model, setting a new benchmark in the field. While previous attempts have been made at half-truth detection, our approach is, to the best of our knowledge, the first to attempt to debunk half-truths.
\end{abstract}

\section{Introduction}
The dissemination of disinformation, especially in the form of half-truths, can have significant and negative implications as it has the potential to disrupt social and economic harmony \cite{NBERw23089, Su2020}. A recent example of this was seen during the Covid-19 vaccination drive, where the spread of disinformation led to widespread fear and skepticism among the public regarding the efficacy and safety of the vaccine~\cite{he2021racism, DBLP:journals/corr/abs-2006-11343}.

Our work tackles half-truths by utilizing the LIAR-PLUS dataset \cite{alhindi-etal-2018-evidence} for half-truth detection. There are many forms of half-truth such as deception, exaggeration, propaganda, and intentionally hidden facts, etc. In this work, we only deal with half-truths related to deception and intentionally hidden facts. To improve upon the LIAR-PLUS dataset, we added a new column to it, called \textit{\textbf{shortened justification}}, using the concept of textual entailment. This shortened justification is referred to as \textbf{\textit{support}} when the label is true or mostly-true, and \textbf{\textit{counter}} when the label is half-true, false, barely-true, or pants-on-fire. \textit{Supports} or \textit{counters}, in our context, are explanations for the label associated with each claim. We refer to these explanations as \textbf{\textit{evidence}} in our work. Our approach not only detects half-truths but also aims to debunk the claim by editing and transforming it into a truthful statement. \textbf{\textit{‘Claim’}}, as coined by \cite{toulmin_2003}, is \textit{‘an
assertion that deserves our attention’}. In our study, a \textit{claim} is defined as a textual statement that can be made by individuals, news websites, political parties, and other sources.

This research is a significant advancement in the field of natural language processing (NLP) and has the potential to contribute to fact-checking and computational journalism, ultimately helping to prevent people from falling prey to disinformation.

\begin{figure*}
    \includegraphics[width = 15cm]{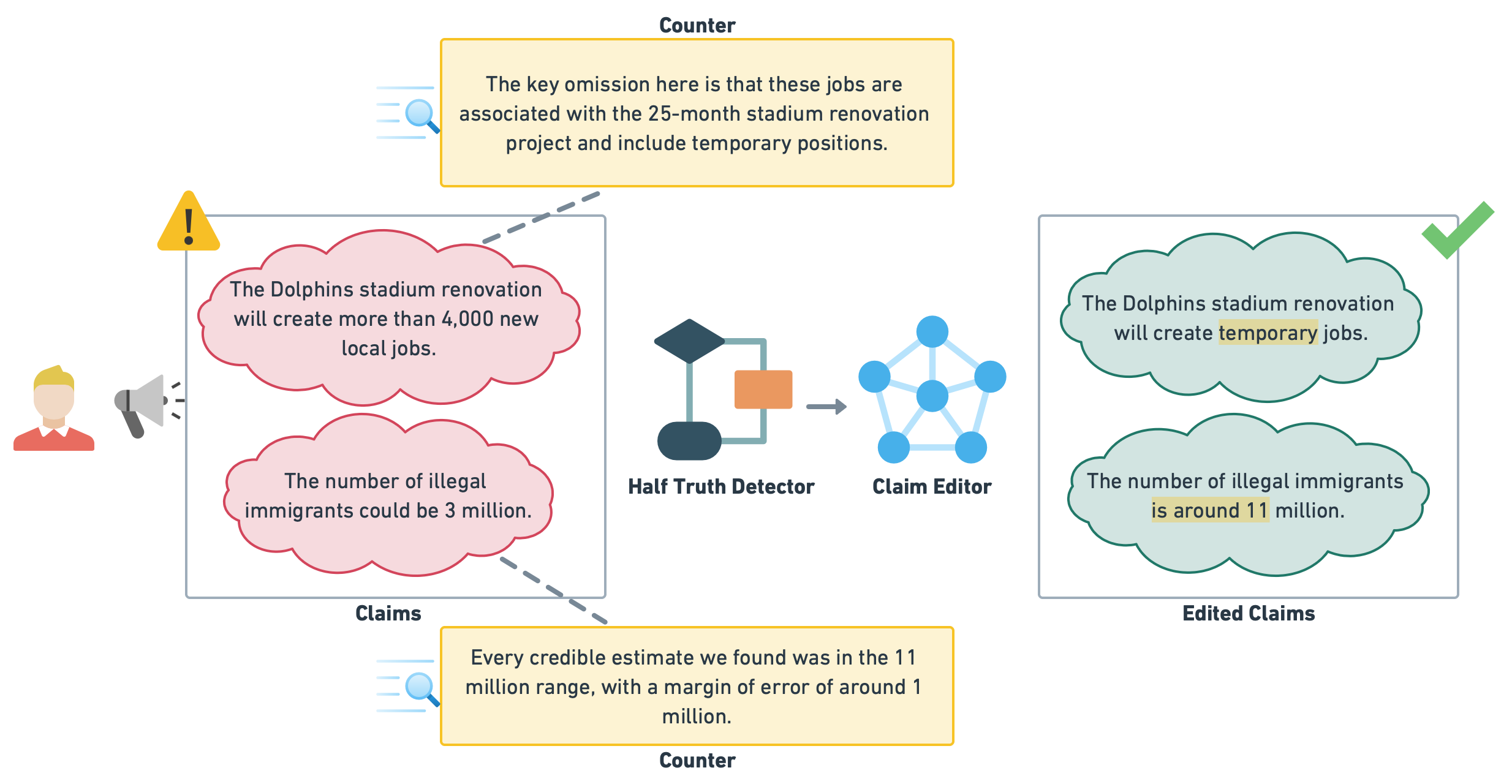}
    \caption{Picture depicting the half-truth detection and debunking pipeline}
    \label{fig:half-truth-pipeline}
\end{figure*}

A \textbf{\textit{half-truth}} is a statement that is partially true but intentionally omits important details that would significantly alter its meaning. This type of statement is deceptive as it can lead to misunderstandings or false impressions. Even if a statement is technically true, it cannot be considered entirely truthful if it excludes crucial information. Half-truths are lies of omission.

For instance, an example of a half-truth is the statement \textit{"Electronic gadgets mandatory for e-census in 2023"}, which contains a hidden piece of information that people need not buy the gadgets since the government will provide them. The statement is partially accurate, but it is also misleading because it fails to disclose a crucial detail that could cause confusion for the reader. It is important to include all necessary information to ensure clarity and prevent misunderstandings.

\subsection{Motivation}
Generating sensational or misleading content can attract more viewership, engagement, and advertising revenue. Some individuals and groups exploit this for personal gain, including political, financial, or ideological motivations. However, traditional fact-checking methods rely on human fact-checkers and can be time-consuming \cite{10.1145/2806416.2806652}, which limits their effectiveness in responding to the constant stream of disinformation. This is where automated fact-checking \cite{guo-etal-2022-survey} and disinformation debunking systems become crucial, as they can quickly detect \cite{DBLP:journals/corr/abs-1902-06673} and respond to disinformation in real-time, which can help limit its reach \cite{37381}. Our detection and debunking pipeline is a powerful tool that can help detect half-truths and false claims faster, thus saving valuable person-hours that would otherwise be spent on manual fact-checking. Additionally, our pipeline is equipped with an evidence extraction tool that allows us to collect evidence faster, which can then be used to debunk half-truths and false claims faster. Overall, our system is an important step toward helping combat the spread of disinformation on digital platforms.

\subsection{Problem Definition}
In this work, we have implemented a half-truth detection model to detect half-truths. Given a claim \textbf{$C$} and the corresponding evidence \textbf{$E$} as input, the half-truth detection model predicts whether the given claim is true or half-true, or false. It is a three-class classification problem.\\
In addition to that, we have implemented a claim editing model to edit \textit{half-true} and \textit{false} claims. Given a \textit{half-true} or \textit{false} claim \textbf{$C$} and the corresponding evidence \textbf{$E$} as input, our claim editing pipeline uses the evidence to edit the \textit{half-true} or \textit{false} claim and tries to generate an edited \textit{true} claim\textbf{ $C*$} with control over editing the selected parts of input claim. The overall task is depicted in Figure \ref{fig:half-truth-pipeline}.

\subsection{Contributions}
Our contributions are:
\begin{enumerate}
    \item Extending the LIAR-PLUS dataset by adding an extra column called \textit{shortened justification}, using textual entailment, and achieving a benchmark accuracy of \textbf{82\%} for the half-truth detection model.
    \item Devising a novel claim editing technique, which aided the T5 model, to outperform cutting-edge systems such as GPT2 by 82\%, Roberta by 57\%, PEGASUS by 42\%, and Tailor by 23\% with a success rate of\textbf{ 85\%} in the task of debunking claims and achieving an average content preservation score of \textbf{0.88} (on a scale of 0-1) on the edited claims.
\end{enumerate}

\subsection{Roadmap}
The paper is structured as follows: In section \ref{related work}, we have discussed the related works. Section \ref{detection system} presents a model for detecting half-truths, while section \ref{claim editing} proposes a technique for editing half-true claims. The experiments and results are discussed in \ref{exp and results}  and the conclusion and future work are outlined in section \ref{conclusion}.

\section{Related Work}\label{related work}
Automated fact-checking is a process that involves the detection of the veracity of claims by extracting relevant evidence and validating the claim against the evidence. This has become a topic of immense interest and research in recent years, resulting in numerous works in the domain of fact-checking. \citet{alhindi-etal-2018-evidence} introduced the LIAR-PLUS dataset and implemented a veracity prediction model using LSTM. The authors' work involved training the LSTM model to classify statements into six different categories based on their truthfulness, thereby demonstrating the effectiveness of deep learning approaches in this field. FEVER \cite{thorne-etal-2018-fever} is a popular dataset in the domain of fact-checking.

The idea of detecting veracity using the emotion of the claim was proposed in \citet{DBLP:journals/corr/abs-1903-01728}. The authors used sentiment analysis to detect the emotion conveyed by a claim and then used this to infer the claim's veracity. \citet{DBLP:journals/corr/abs-1911-05885} discusses the computational complexity of deception by half-truth. The authors demonstrated that half-truths can be computationally more challenging to detect than other forms of deception, thus emphasizing the need for specialized approaches to identify and address this issue. Building on this idea, \citet{inbook} filtered out half-truths during fake news detection and expressed their idea of detecting half-truths in the future. Motivated by this idea, our work attempts to address the half-truth detection problem.

Along with half-truths, there are other forms of disinformation, such as fake news and exaggerated and sensationalized news. \citet{DBLP:journals/corr/abs-2108-13493} focuses on detecting exaggeration in the claims made by press releases. The authors propose a supervised learning approach that utilizes sentence-level features to detect exaggerated claims. \citet{li-etal-2017-nlp} conducted an analysis and inspection of exaggerated claims in the domain of scientific news. The authors proposed a framework that leverages natural language processing techniques to detect exaggerated claims in scientific news articles.

In addition to detecting half-truths, debunking them is necessary to keep a check on the spread of disinformation. \citet{popat-etal-2018-declare} discusses debunking fake news using external evidence. \citet{atanasova-etal-2020-generating-fact} discusses generating explanations along with detection.  \citet{schuster-etal-2021-get} discusses about robust fact verification using evidence that changes with time. Counterfactuals called contrast sets \cite{DBLP:journals/corr/abs-2004-02709} can be created to debunk half-truths. Moreover, structural-level properties of a claim can be used to edit it by semantically controlling the text generation \cite{DBLP:journals/corr/abs-2107-07150}. Building on these ideas, we have developed a controlled claim editing technique that makes minimum edits to a half-true claim to make it true. Our approach leverages the structural-level properties of a claim to identify the necessary edits and then uses semantically controlled text generation to make the claim true.

Overall, our work attempts to address the challenging problem of half-truth detection and debunking by utilizing a combination of techniques, including contrast sets and controlled claim editing. By detecting and debunking half-truths, we can take a step toward combating disinformation and promoting the spread of accurate information.

\section{Half-truth detection model}\label{detection system}
For half-truth detection, the task is to build a model that takes a claim and evidence and predicts one of the three labels: true, half-true, or false for the given input. Given a claim \textbf{C} and evidence \textbf{E}, and \textbf{y} is the generated label, we can mathematically formulate the task as follows:
\begin{align}
    y^* &= \underset{y \in \{true,false,half-true\}}{argmax} P(y|C;E)
\end{align}

\subsection{LIAR-PLUS-PLUS Dataset}

\begin{figure}

\fbox{\begin{minipage}{19em}
\small
\textbf{Statement:} \textit{“Says Rick Scott cut education to pay for even
more tax breaks for big, powerful, well-connected corporations.”}\\
\textbf{Speaker:} \textit{Florida Democratic Party}\\
\textbf{Context:} \textit{TV Ad}\\
\textbf{Label:} \textit{\textbf{half-true}}\\
\textbf{Extracted Justification:} \textit{A TV ad by the Florida Democratic Party says Scott "cut education to pay for even more
tax breaks for big, powerful, well-connected corporations."
However, the ad exaggerates when it focuses attention
on tax breaks for "big, powerful, well-connected corporations." Some such companies benefited, but so did many
other types of businesses. And the question of whether the
tax cuts and the education cuts had any causal relationship
is murkier than the ad lets on.}\\
\textbf{Shortened Justification:}\textit{ However, the ad exaggerates when it focuses attention on tax breaks for "big, powerful, well-connected corporations." Some such companies benefited, but so did many other types of businesses.}
\caption{An excerpt from the LIAR-PLUS-PLUS dataset}
\label{fig: LIAR-PLUS-PLUS dataset example}
\end{minipage}}
\end{figure}

The LIAR-PLUS dataset \cite{alhindi-etal-2018-evidence} is a benchmark dataset for fact-checking research, extended from the LIAR dataset \cite{wang-2017-liar}. It has 12,000+ human-labeled statements classified into six categories: \textit{true}, \textit{mostly-true}, \textit{half-true}, \textit{barely-true}, \textit{false}, or \textit{pants-on-fire}. The dataset was collected from the Politifact\footnote{\href{https://www.politifact.com}{Politifact: Website}}. We extended LIAR-PLUS by adding the \textbf{\textit{shortened justification}} column and called it \textbf{LIAR-PLUS-PLUS}. This column was created by using textual entailment on the extracted justification column of the LIAR-PLUS dataset. An excerpt from the LIAR-PLUS-PLUS dataset is presented in Fig \ref{fig: LIAR-PLUS-PLUS dataset example}. The composition of LIAR-PLUS-PLUS data for training, validation, and test split is 10240, 1284, and 1283 instances respectively. The average number of sentences in the extracted justification is 6. The shortened justification is at max 2 sentences and a minimum of one sentence.

\subsubsection{Creation of shortened justification}\label{shortened-justification}
To extract shortened justifications, we utilized a natural language inference (NLI) model that assigns entailment scores to pairs of sentences. Our algorithm employs this NLI model to generate supports and counters for each claim in the LIAR-PLUS dataset. A support is a statement that strengthens the claim, while a counter is a statement that challenges it. This is accomplished by calculating the entailment scores between each sentence in the claim and its corresponding justification. Each sentence is classified into one of three labels: entailment, contradiction, or neutral.

The rationale behind employing textual entailment is as follows. True and mostly-true claims typically have supporting text, which aligns with the \textit{entailment} label. False and pants-fire claims, on the other hand, tend to have text that contradicts them, similar to a \textit{contradiction} label. Half-true and barely-true claims often contain text that mentions hidden information or the deceptive aspect of the claim, which cannot be directly entailed or contradicted and thus corresponds to a \textit{neutral} label. With this idea, we can have sufficient information in the evidence to detect these labels.

In our approach, we calculate the entailment scores on a sentence-by-sentence basis between the claim and each piece of evidence. Among all the sentences predicted as \textit{entailment} or \textit{contradiction}, we select the sentence with the highest confidence or probability score as the first part of the shortened justification. For sentences predicted as \textit{neutral}, we choose the sentence with the highest confidence as the second part of the shortened justification. If no sentences are predicted as \textit{neutral}, we have only one sentence in the shortened justification, and vice versa. As a result, the shortened-justification consists of a maximum of two sentences and a minimum of one sentence. We refer to this shortened justification as evidence, which serves as an explanation for the label associated with each claim. The details of our NLI model are discussed in section \ref{NLI model} of the appendix.

We have performed a manual evaluation of the extracted shortened-justification using 2 evaluators. We selected 100 claims from each label and asked the annotators to check if the shortened justification contains the required information to predict the label of the claim. Out of 600 claims, for 568 claims, the annotators found sufficient information in the shortened-justification. This amounts to around \textbf{94.6\%} of successful extraction of evidence.

We created a network with BERT (\cite{DBLP:journals/corr/abs-1810-04805} based classifier model. We call this model, the half-truth detection model. We used the LIAR-PLUS dataset and the LIAR-PLUS-PLUS dataset separately to train different versions of the model. The two versions are called \textbf{J(model trained on LIAR-PLUS dataset)} and \textbf{SJ (model trained on LIAR-PLUS-PLUS dataset)}. For version J, the input is claim and justification, and for version SJ, the input is the claim and shortened justification. We trained the models for the task of tri-class classification; true, false, and half-truth by grouping claims with half-true and barely-true labels into half-true, true, and mostly-true labels as true and false and pants-on-fire as false. This grouping has been done based on the annotation policy of the Politifact website and truth-o-meter\footnote{\href{https://www.politifact.com/article/2018/feb/12/principles-truth-o-meter-politifacts-methodology-i/}{The Principles of Truth-O-Meter: Webpage}} definitions. According to the definitions of truth-o-meter, barely-true statements are claims that have hidden information. Since, this falls into the category of half-truths, we have grouped barely-true and half-true claims into a single label. The composition of the LIAR-PLUS-PLUS dataset after grouping the labels is shown in Table \ref{tab:LIAR++ composition}.

\begin{table}
      \centering
      \begin{tabular}{|l|*5{c|}}\hline
      Label & Train & Test & Validation\\\hline
      true & 3649 & 460 &420\\
      half-true & 3780 & 481 &485\\
      false & 2840 & 342 &379\\\hline
      \end{tabular}
      \caption{The composition of train, test, and validation split of the LIAR-PLUS-PLUS dataset after grouping the labels.}
      \label{tab:LIAR++ composition}
\end{table}

We tested both models on the test set of the LIAR-PLUS-PLUS dataset. We obtained an F1 score of\textbf{ 0.82}, which is considered a benchmark accuracy, for the SJ version of our model. On the other hand, for the J version of the model, we achieved an F1 score of\textbf{ 0.724}. The SJ version of the model outperformed the J version because it was presented with only the pertinent information in the justification. Therefore, we can confidently assert that the textual entailment task assisted in enhancing the accuracy of the half-truth detection model. By extracting solely the relevant information from the justification, we have successfully reduced the complexity of the model. The model is required to examine the shortened justification to make a determination. Internally, the model is inferring if the statement is entailing or contradicting the shortened justification. As a result, the model can discern that half-truths are lies of omission, whereas truthful news entails and fake news contradicts the shortened justification. Logistic Regression and SVM models were used as baseline models, and they were trained on GloVe embeddings \cite{pennington-etal-2014-glove} using both datasets (after grouping labels). The F1 scores of all these models are presented in Table \ref{tab:Accuracy scores}. All these models are tested on the test set of the LIAR-PLUS-PLUS dataset. The confusion matrix and per-label macro averages of the best-performing model, BERT-based sequence classifier, trained using the LIAR-PLUS-PLUS dataset is presented in the Tabel \ref{tab:Confusion matrix}. The macro average precision, recall, and F1 scores are \textbf{0.811}, \textbf{0.831}, and \textbf{0.82} respectively.

\begin{table}
      \centering
      \begin{tabular}{l r c c c}
        \toprule
        \textbf{Model} & \textbf{Data} & \textbf{F1} \\
        \midrule
        SVM & J & 67.5 \\
        LR & J & 66.8 \\
        BERT & J & 72 \\
        \midrule
         SVM & SJ & 72.6 \\
        LR & SJ & 71.4 \\
        BERT & SJ & \textbf{82} \\
        \bottomrule
      \end{tabular}
      \caption{
      Performance of Half-Truth detection module with various model-data combinations. Abbreviations {F1}:= F1-accuracy score \%, {J}:= Model trained using the LIAR-PLUS dataset, {SJ}:= Model trained using the LIAR-PLUS-PLUS dataset, {LR}:= Logistic Regression Classifier.
      }
      \label{tab:Accuracy scores}
\end{table}

\begin{table*}
      \centering
      \begin{tabular}{|l|*{6}{c|}}\hline
      & & \multicolumn{3}{| c |}{\textbf{Gold Labels}}&\\\hline
      & & \textbf{true} & \textbf{false} & \textbf{half-true} & \textbf{per-label precision}\\\hline
      & \textbf{true} & 364 & 0 & 96 & 0.791\\
\textbf{Model Outputs} & \textbf{false} & 2 & 272 & 68 & 0.795\\ 
& \textbf{half-true} & 31 & 42 & 408 & \textbf{0.848}\\\hline
& \textbf{per-label recall} & \textbf{0.916} & 0.866 & 0.713 & \\\hline    
      \end{tabular}
      \caption{Confusion matrix and per-label precision and recall of the best-performing model, BERT-based sequence classifier, trained using the LIAR-PLUS-PLUS dataset }
      \label{tab:Confusion matrix}
\end{table*}

\section{Claim Editing Model}\label{claim editing}

The claim editing model is useful to edit a half-true or false claim. We can perform controlled claim editing using our technique. By control, we mean, we can make precise adjustments to the selected parts of the claim. We mask the tokens that need replacement and use the objective of masked language modeling to fill those masks. Thus, an edited claim is generated. Our objective is to convert false or half-true claims into true claims. If we find the contradicting part or the deceptive part in the claim and mask it, we can replace it with the correct part from the evidence. By making this adjustment to the false and half-true claims, they can be converted to true claims. So, firstly, we need a model to fill masked tokens using the context provided to the model.

\subsection{TAPACO Dataset}
We utilized the TAPACO paraphrase dataset \cite{scherrer-2020-tapaco} and augmented it. The TAPACO dataset is a paraphrase corpus consisting of 73 languages that were extracted from Tatoeba\footnote{\href{https://tatoeba.org/en/}{Tatoeba: Website}}, which is a crowdsourcing project primarily for language learners. We have selected 60,000 instances from the TAPACO dataset to augment our dataset, where each instance consists of an original sentence and its corresponding paraphrased sentence.
\\
\textbf{Original sentence:} \textit{Many people respect you. Do not disappoint them.}
\\
\textbf{Paraphrased sentence:}\textit{ A lot of people look up to you. Do not let them down.}
\subsubsection{Dataset Augmentation:}\label{augmented dataset}
We have obtained Semantic Role Labeling (SRL) tags for all paraphrased sentences and appended an additional column to the TAPACO dataset. Our approach involved utilizing the SRL generator provided by Allen AI \footnote{\href{https://allenai.org/}{Allen AI: Website}} to extract SRL tags for each sentence. In some cases, the SRL generator produced multiple outputs for a single paraphrased sentence. We selected the output that contained the highest number of semantic roles (tags) to resolve this issue. The augmented dataset, which now contains three columns, as shown below, was utilized to train the claim editing model.
\\
% \textbf{Example:}
% \\
\textbf{Original sentence:} \textit{Many people respect you. Do not disappoint them.}
\\
\textbf{Paraphrased sentence:}\textit{ A lot of people look up to you. Do not let them down.}
\\
\textbf{SRL tagged Paraphrased sentence:} \textit{[ARG0: A lot of people] [V: look] [ARG1: up to you] . Don't let them down.}

We use the augmented TAPACO dataset to train the T5 model for the task of claim editing. T5 is a text-text transformer model from Google. T5 is used for a variety of purposes, including machine translation, summarization, and masked language modeling. We took 50000 instances from the augmented dataset as a training split with a validation and test split of 5000 instances each. We have given the SRL tags of the paraphrased sentence and the original sentence with a few masked tokens as input to T5 and expect the original sentence as output.
\vspace{2mm}
\\
\textbf{Input:} \textit{[[ARG0: A lot of people] [V: look] [ARG1: up to you] . Don't let them down .] Many people extra\_id\_0\ you. Don't extra\_id\_1 them. }
\vspace{2mm}
\\
\textbf{Output:} \textit{Many people respect you. Don't disappoint them.}
\vspace{2mm}
\\
During training, we have masked only nouns, adjectives, and verbs in the original claim to make sure that the model is not only learning the structural properties of the language but also semantics. That is the main motivation for us to use SRL-tagged paraphrased sentences in the input. This idea of using SRL tags to perturb and edit sentences have been used by \citet{ross-etal-2022-tailor}. With this mechanism of training, the model learns to fill the masked tokens by using the SRL-tagged paraphrased sentence. Hence, it maximizes the context (the paraphrased sentence) to fill only the masked tokens thereby minimizing the reconstruction loss. Now that our objective of filling masked tokens with the SRL-tagged context provided in the header (enclosed in []) is achieved by T5, we can adapt this to our task of claim editing.

For the task of claim editing, we can provide the SRL-tagged evidence in the header and the masked half-true or false claim as input to the T5 model. The model will fill the masked tokens using the SRL-tagged evidence. For editing half-true and false claims in the LIAR-PLUS-PLUS dataset, we use the 'shortened justification' as evidence. We call it the counter.
We used the Allen AI SRL tag generator to extract SRL tags of counters. If we have multiple outputs from the SRL generator, we take the one with the maximum number of tags as the SRL-tagged counter. In addition to that, we need to mask the deceptive and contradicting part of the claim. We use the below strategy to mask a false or half-true claim.\\
\textbf{Masking false or half-true claim:}
\\
To ensure accurate editing of the claim, it is necessary to mask specific tokens rather than selecting them randomly. To identify the appropriate tokens for masking, we employ the concepts of textual entailment and cosine similarity. Using an AllenNLP constituency parser, we divide the claim into multiple segments. Only those segments that contradict the evidence or exhibit lower similarity are considered for replacement with a mask. By utilizing an NLI model, we calculate scores indicating the degree of contradiction. If no contradictory segments are found, we mask the segment that exhibits lower similarity with the counter. This strategy facilitates the T5 model in accurately filling the masked tokens using the counter, thereby ensuring precise editing of the relevant tokens.

The T5 model trained by us is capable of generating claims where the number of edits will be less and original content is preserved. We have given the SRL-tagged counter (as header) and the masked claim as input to the T5 model. The model generates a list of edited claims. In our case, we have limited the generations to five in number. We used Constrained Beam Search for text generation. Unlike ordinary beam search, constrained beam search allowed us to exert control over the output of text generation.
\vspace{2mm}
\\
\textbf{Example of claim editing:}
\\
\textbf{Original claim-} \textit{The Dolphins stadium renovation will create more than 4,000 new local jobs.}
\\
\textbf{Masked claim-} \textit{The Dolphins stadium renovation will create extra\_id\_0.}\
\\
\textbf{Counter-} \textit{The key omission here is that these are jobs associated with the 25-month stadium renovation project and include temporary positions.}
\\
\textbf{Input to T5-} \textit{[The key omission here is that [ARG2: these] are jobs associated with the 25-month stadium renovation project and [V: include] [ARG1: temporary positions] .] The Dolphins stadium renovation will create extra\_id\_0.}
\\
\textbf{Output from T5-} List of edited claims (For example, one of the edited claims is \textit{"The Dolphins stadium renovation will create temporary jobs."})
\vspace{2mm}
\\
\textbf{Claim Filtering:}
\\
One edited claim among the list of edited claims will be filtered out as the best claim. We employed a filtering technique that filters out the best-edited claim using the reward mechanism of a claim being true, maximizing the word overlap between the original claim and the edited claim. For the reward mechanism, we have used the SJ version of the half-truth detection model (0.82 F1 accuracy). For a set of edited claims, we find the label predicted by the half-truth detection model. For this model, we give the edited claim and the counter as input. The edited claims that are predicted as true by the half-truth detection model are rewarded highly. Because, now after editing, we have a validation that the half-true or false claims have been converted to true. If none of the edited claims is true, we filter all the edited claims to the next stage. These filtered claims are checked for maximum word overlap with the original claim and the one with the highest overlap is considered the best-edited claim. We give importance to word overlap because we want to edit the claim minimally. The motivation behind this minimal edit was taken from \citet{DBLP:journals/corr/abs-2004-02709}.

\section{Experiments and Results}\label{exp and results}
In this section, we have presented the experiments performed on the task of claim editing task and our findings.
\subsection{Baselines}
We have used different language models for the task of claim editing to compare these results with our technique. We compare our claim editing technique with various language models such as GPT2 \cite{Radford2018ImprovingLU}, RoBERTa \cite{DBLP:journals/corr/abs-1907-11692}, PEGASUS \cite{zhang2019pegasus} and Tailor \cite{ross-etal-2022-tailor}. Please refer to section \ref{claim-editing-lms} of the appendix for the usage of LMs for claim editing and an example.

\subsection{Evaluation Metrics}
We have evaluated the edited claims on two evaluation metrics to make sure the quality of the edits is not compromised in our technique. The two metrics are \textbf{content preservation} and \textbf{disinfo-debunk} (disinformation debunk). A minimally edited sentence must maintain its content. In this paper, \textit{content preservation} was evaluated using the BLEU score \cite{papineni-etal-2002-bleu}. Content preservation is not a similarity metric, it is a metric to measure overlap. The \textit{content preservation} score is the BLEU score between the edited claim and the original claim. This metric measures how overlapped the edited claim and original claim are since we wanted to edit the claim minimally. The end goal of claim editing is to debunk false and half-true claims. Hence, we should evaluate how many claims have been converted to true after editing them using our technique and also the other language models. The \textit{disinfo-debunk} metric measures the percentage of claims that have been converted to true after editing them. We have used the BERT(SJ) model to detect the label of the edited claim and compute the \textit{disinfo-debunk} metric. The computation of the \textit{disinfo-debunk} metric is limited by the accuracy of the BERT(SJ) model in predicting the labels.

\subsection{Evaluation}
We have used the LIAR-PLUS-PLUS dataset as our evaluation dataset. We have edited the half-true and false claims after grouping the six labels into three labels. We have 7433 half-true and false claims in total. 
\subsubsection{Quantitative}
 Our technique has the highest average\textbf{ content preservation} score (BLEU score) of \textbf{0.88} (on a scale of 0-1). The average content preservation score is the average of BLEU scores between all the edited claims and the corresponding original claim. We found that our technique can debunk half-true and false claims with a \textbf{disinfo-debunk} score of \textbf{85\%}. Clearly, our technique outscores cutting-edge systems such as GPT2 by 82\%, Roberta by 57\%, PEGASUS by 42\%, and Tailor by 23\%. Since we used the BERT (SJ) model with 85.8\% accuracy, to compute the disinfo-debunk score, we can credibly claim that our debunking technique works better than Tailor and other language models. Out of 7433 claims, we obtained edited claims for 6844 claims using Tailor. The reason is that Tailor has a perplexity cutoff threshold. If the perplexity of the generated claim is lesser than this cutoff, we don't get any output from Tailor. Hence, for Tailor, the metrics are computed on 6844 claims. The results can be found in Table \ref{tab:State-of-the-art Metric scores}.

\begin{table}
      \centering
      \begin{tabular}{l r r c}
        \toprule
        \textbf{Model} & \textbf{CP} & \textbf{Disinfo-debunk} \\
        \midrule
        Tailor & 0.76  & 4243 / 6844 (62\%) \\
        GPT2 & 0.52  &  223 / 7433 (3\%) \\
        RoBERTa &  0.82  & 2081 / 7433 (28\%) \\
        PEGASUS &  0.16  & 3196 / 7433 (43\%) \\
        Our Technique &  \textbf{0.88} & 6318 / 7433 \textbf{(85\%)} \\
        \bottomrule
      \end{tabular}
      \caption{Evaluation of Language models vs. our technique on the LIAR-PLUS-PLUS dataset for the task of claim editing. Abbreviations {CP}:= Content Preservation}
      \label{tab:State-of-the-art Metric scores}
\end{table}
\subsubsection{Qualitative}
In addition to the qualitative evaluation of edited claims, we performed the human evaluation of the edited claims on two metrics, fluency, and edit-correctness.\\
\textbf{Fluency} (On a scale of 1-3):\\
Less fluent and incorrect grammar- score of 1\\
Medium level fluency- score of 2\\
Fluent and grammatically correct- score of 3\\
\textbf{Edit-correctness} (On a scale of 1-3):\\
Incorrect edit- score of 1 (Edited the incorrect part)\\
Partially correct edit- score of 2 (Edited the right part but not correctly edited)\\
Correctly edited- score of 3\\
We have manually annotated randomly picked 100 claims edited by the T5 model. Two annotators annotated for fluency and edit-correctness. The inter-annotator agreement, Cohen's Kappa (\cite{10.1162/coli.07-034-R2}, was found to be \textbf{74.86} for fluency and \textbf{66.28} for edit-correctness. The average fluency was \textbf{2.75} and the average edit-correctness is \textbf{2.3}. The averages are computed on annotated 100 claims by both annotators.

We also wanted to compare our technique with Tailor. We randomly picked 50 instances and the corresponding edited claims edited using Tailor and T5. We did not mention which claim is edited by which model. Hence, we avoid biased responses. We shared the annotation guideline with the \textbf{4 users} and asked them to rate the edited claims for two metrics, fluency and edit-correctness. For claims edited by the T5 model, we found the average fluency was \textbf{2.68} and the average edit-correctness was \textbf{2.42}. For claims edited by Tailor, we found the average fluency was \textbf{2.28}  and the average edit-correctness was \textbf{1.36}. Our technique performs better than Tailor in human evaluation too.

\section{Conclusion and Future work}\label{conclusion}

Our study demonstrates that T5 surpasses sophisticated techniques like Tailor in effectively debunking half-truths through controlled claim editing. T5's ability to accurately fill in the necessary information with minimal edits contributes to its superior performance. Accurate detection of half-truths can be achieved with good accuracy, aided by the creation of a shortened justification column. The utilization of textual entailment and SRL-tagged evidence highlights the significance of NLP models in understanding linguistic properties.

Moving forward, our future plans involve creating and annotating a dataset specifically for gathering a larger quantity of high-quality data on half-truths from news articles. We aim to develop a novel algorithm using reinforcement learning to select and rank phrases within a claim for editing purposes. Additionally, we aspire to build a reliable and trustworthy real-time evidence extraction module to facilitate the detection and debunking of disinformation. While we have successfully debunked half-true and false claims using evidence from the LIAR-PLUS-PLUS dataset, the same cannot be said for real-time half-truths. To address this, we have developed a real-time evidence extraction module that retrieves results from Google News and extracts article summaries. Summaries from reputable sources can serve as valuable evidence. However, further large-scale testing is still pending.

\section{Limitations}\label{limitations}

One of the major challenges that we have come across is the lack of adequate data in the LIAR-PLUS dataset. The NLI model used for creating the shortened justification column in the LIAR-PLUS-PLUS dataset is not 100\% accurate. Hence, the newly created column needs manual inspection and correction. The masking algorithm needs improvement since we are using only the NLI model and similarity score to mask claims.

\section*{Acknowledgements}
We thank the Politifact website and its annotators for their hard work in annotating claims. We also thank the annotators who participated in the experiments and submitted their valuable responses.

\section*{Ethics Statement}
All the annotators in the experimental study gave their consent on submitting their responses. The annotation guideline was provided to them. There was anonymity maintained in the collection of responses in Google Forms. No personal data was collected during our experiments.

\bibliography{anthology,custom}
\bibliographystyle{acl_natbib}

\appendix

\section{Appendix}\label{sec:appendix}
This section presents detailed information about using LMs for the task of claim editing in section \ref{claim editing}.
\subsection{NLI model}\label{NLI model}
We have used the NLI (Natural Language Inference) model for extending the LIAR-PLUS dataset. For the details related to the usage of this model, refer to section \ref{shortened-justification}. We have trained a BERT-based NLI model using SNLI \cite{DBLP:journals/corr/abs-1805-02266} and MNLI \cite{DBLP:journals/corr/WilliamsNB17} datasets. 
We took a combination of 300,000 instances from both datasets as a training split to train the NLI model for the task of producing entailment scores with validation and a test split of 5000 instances each. The average F1 score is 0.91. The model can predict the textual entailment label with 91\% accuracy. The label-wise F1 scores for entailment, contradiction, and neutral are 93\%, 92\%, and 87\% respectively.

\subsection{Claim editing using multiple Language models}\label{claim-editing-lms}
\subsubsection{GPT2}
We have provided a three-fourths length of the claim as a prompt to the GPT2 model and the generated output from GPT2 is considered the edited claim. We consider the first sentence generated by GPT2 as the edited claim.

\subsubsection{RoBERTa}
We have used the RoBERTa-based transformer model from Hugging Face for the task of text infilling. We have provided the masked claim, provided to the T5 model, concatenated with the counter as input. The masked positions of the claim will be filled by this model and the filled claim is considered the edited claim.

\subsubsection{PEGASUS}
We have used the PEGASUS model for generating summaries. We provide the extracted justification from LIAR-PLUS-PLUS as input to the model and the generated summary is considered the edited claim. Since the extracted justification contains the explanation for every label, it is fair to consider the summary as a true claim.

\subsubsection{Tailor}
We have used Tailor \cite{ross-etal-2022-tailor} for perturbing the claim by maximizing the counter. We have used the perturb\_with\_context function from Tailor. This function is used to fill the masked part of the claim using the counter. The inputs for the function are, the claim, part of the claim which needs to be edited, and the counter. The perturbed claim is considered the edited claim.

\fbox{\begin{minipage}{19em}
\small
\textbf{Claim:} The number of illegal immigrants could be 3 million.\\
\textbf{Counter:} Every credible estimate we found was in the 11 million range, with a margin of error of around 1 million.\\
\textbf{Masked Claim:} The number of illegal immigrants <extra\_id\_0> <extra\_id\_1>.\\
\textbf{SRL-tagged counter:} [ARG1: Every credible estimate we found] [V: was] [ARG2: in the 11 million range], [ARGM-ADV: with a margin of error of around 1 million].\\
\textbf{Extracted Justification:} Trump said the number of illegal immigrants "could be 3 million. It could be 30 million. "Both figures are not within the range of possibility. Every credible estimate we found was in the 11 million range, with a margin of error of around 1 million. The figure has "always" been 11 million, in recent years, because of the flow of undocumented immigrants in and out of the United States.\\
\textbf{Tailor:}\\
\textbf{Input:} SRL-tagged counter + Masked Claim\\
\textbf{Output:} The number of illegal immigrants estimated at 11 million.\\
\textbf{GPT-2:}\\
\textbf{Input:} The number of illegal immigrants\\
\textbf{Output:} The number of illegal immigrants could be even more alarming, given many of them came to the United States just to work legally in one of the most dangerous areas on earth.\\
\textbf{RoBERTa:}\\
\textbf{Input:} Counter + Masked Claim\\
\textbf{Output:} The number of illegal immigrants could be trumped.\\
\textbf{PEGASUS:}\\
\textbf{Input:} Extracted Justification\\
\textbf{Output:} The number of illegal immigrants in the United States is "not within the range of possibility", according to Republican presidential candidate Donald Trump.\\
\textbf{Tailor:}\\
We call a function \textbf{\textit{perturb\_with\_context}} with Claim, masked part of the claim, and counter.\\
\textbf{Output:} The number of illegal immigrants could be in the 11 million range, with a margin of error of around 1 million.
\end{minipage}}

\end{document}